%% file: main.tex
\documentclass{tdp}

\input{imports.tex}

\begin{document}

\title{A Hierarchical Approach for Assessing the Vulnerability of Tree-Based Classification Models to Membership Inference Attack}

\author{Richard~J.~Preen$^{}$, Jim~Smith$^{}$}

\address{$^{}$%
Department of Computer Science and Creative Technologies,
University of the West of England, Bristol, UK\@.\\ 
E-mail: {\small \tt{\{richard2.preen,james.smith\}@uwe.ac.uk}}
}

%\thanks{X} 

\TDPRunningAuthors{Richard~J.~Preen, Jim~Smith}
\TDPRunningTitle{A Hierarchical Approach for Assessing the Vulnerability of Tree-Based Models}
\TDPThisVolume{X}
\TDPThisYear{2025}
\TDPFirstPageNumber{1}
%\TDPSubmissionDates{Received 13 February 2025\@; received in revised form 12 June 2025\@; accepted XXXX}

\maketitle

\input{00_abstract.tex}


\input{01_introduction.tex}
\input{02_background.tex}
\input{03_methodology.tex}
\input{04_results.tex}
\input{05_conclusion.tex} 
\input{06_acknowledgement.tex}

\bibliographystyle{abbrv}
\bibliography{references}

\input{07_appendix.tex}

\end{document}

%% file: imports.tex
\usepackage[utf8]{inputenc}
\usepackage[T1]{fontenc}
\usepackage{amsfonts}
\usepackage{amsmath}
\usepackage{amssymb}
\usepackage{booktabs}
\usepackage{graphicx}
\usepackage{subfig}
\usepackage{nicefrac}
\usepackage{microtype}
\usepackage{url}
\usepackage{xcolor}
\usepackage{multirow}
\usepackage{caption}
\usepackage{tabularx}
\usepackage{colortbl}
\usepackage{pgfplots}
\usepackage{stackengine}
\usepackage{hyperref}

\pgfplotsset{compat=1.17}
\newcommand{\cellcolorgrey}[1]{ % Map cell to grey shading
    \begingroup
    \edef\tempa{#1}
    \ifx\tempa\empty%
    \else
        \pgfmathsetmacro{\scaled}{max(0,min(100,round(100*abs(#1))))}
        \xdef\clr{gray!\scaled!white}
        \cellcolor{\clr}{#1}
    \fi
    \endgroup
}

\DeclareCaptionLabelFormat{bold}{\textbf{#1~#2}}
\graphicspath{{figs/}}
\makeatletter
\def\input@path{{figs/}}
\makeatother

%% file: 00_abstract.tex
\begin{abstract}
    Machine learning models can inadvertently expose confidential properties of their training data, making them vulnerable to membership inference attacks (MIA). While numerous evaluation methods exist, many require computationally expensive processes, such as training multiple shadow models. This article presents two new complementary approaches for efficiently identifying vulnerable tree-based models: an \textit{ante-hoc} analysis of hyperparameter choices and a \textit{post-hoc} examination of trained model structure. While these new methods cannot certify whether a model is safe from MIA, they provide practitioners with a means to significantly reduce the number of models that need to undergo expensive MIA assessment through a hierarchical filtering approach.

More specifically, it is shown that the rank order of disclosure risk for different hyperparameter combinations remains consistent across a range of datasets, enabling the development of simple, human-interpretable rules for identifying relatively high-risk models before training. While this \textit{ante-hoc} analysis cannot determine absolute safety since this also depends on the specific dataset, it allows the elimination of unnecessarily risky configurations during hyperparameter tuning. Additionally, computationally inexpensive structural metrics serve as indicators of MIA vulnerability, providing a second filtering stage to identify risky models after training but before conducting expensive attacks. Empirical results show that hyperparameter-based risk prediction rules can achieve high accuracy in predicting the most at risk combinations of hyperparameters across different tree-based model types, while requiring no model training. Moreover, target model accuracy is not seen to correlate with privacy risk, suggesting opportunities to optimise model configurations for both performance and privacy.
\end{abstract}

%% file: 01_introduction.tex
\section{Introduction}%
\label{sec:introduction}

An increasing body of work has shown that machine learning (ML) models may expose confidential properties of the data on which they are trained~\cite{Rigaki:2023}. This has resulted in a wide range of proposed attack methods with varying assumptions that exploit the model structure and/or behaviour to infer sensitive information. For example, to perform data reconstruction, model extraction, attribute inference, and membership inference~\cite{Hu:2022, Jegorova:2023}.

The General Data Protection Regulation (GDPR) (and other laws, according to the data and jurisdiction) mean that before an ML model trained on confidential data can be released into the public domain, its vulnerability to attack should be assessed~\cite{Vaele:2018}. One of the most fundamental techniques to assess ML privacy disclosure is the membership inference attack (MIA)~\cite{Shokri:2017}, which attempts to identify whether an individual record was part of the original training data. Since many of the other attacks have been shown to be more difficult to perform~\cite{Zhao:2021}, vulnerability to MIA can be seen as a strong indicator of potential privacy risks, and its presence suggests that further, more sophisticated attacks might also be feasible, even if more challenging to execute. Therefore, a model's robustness against MIA is a crucial first step in demonstrating its privacy-preserving capabilities before public deployment.

However, some of the currently most effective MIAs require considerable computational effort. For example, running a Likelihood Ratio Attack (LiRA)~\cite{Carlini:2022} typically involves sampling (and hence training) 50--100 shadow models. Therefore, from the perspective of people training models, and those tasked with that risk assessment, there is significant value in being able to rapidly identify vulnerable models or reduce the number that require computationally expensive evaluation. 

In this paper we investigate two approaches to this, which also cast light on the closely related issue of what are the underlying causes of model vulnerability, and what can be measured as proxies for those. First we explore the relationship between model hyperparameters and risk. That is, we use an empirical approach to learn simple human-interpretable rules that identify whether a model is \textit{relatively} high risk based on the hyperparameters independent of the training dataset and before any model fitting. These rules may provide additional insight into the causes of disclosure risk and enable researchers to avoid training unnecessarily high risk models, e.g., such combinations could be removed from a typical hyperparameter search. Furthermore, they may provide additional information to support making an assessment of whether a model should be publicly released.

However, since the risk of information disclosure depends on characteristics of both model and dataset, ascertaining the \textit{absolute} risk posed by a model must be performed after training. Therefore in the second strand of work we leverage decades of statistical disclosure control (SDC) practice for risk-assessing tables and regression models to design a number of computationally cheap structural measures based on the trained model and how it partitions the training data.

The contributions of this paper are:

\begin{enumerate}
    \item{In terms of \textit{ante-hoc} analysis of hyperparameter choices, we show that:}
        \begin{enumerate}
            \item The rank order of risk for different hyperparameter combinations is preserved across different datasets, i.e., is dataset independent.
            \item Risk is not correlated with target accuracy. 
        \end{enumerate}
        Therefore we can identify a region of unnecessary risk---which can be characterised by simple human-readable models, and used to help researchers avoid creating risky models in advance.
    \item In terms of \textit{post-hoc} analysis of trained model structure and how it partitions the decision space we show that:
        \begin{enumerate}
            \item We can design useful cheap-to-compute structural metrics based on properties of target models and their behaviour with training sets. 
            \item Empirical results suggest a combination of these structural metrics constitutes a sufficient but not necessary indicator of whether a target model is vulnerable to MIA attacks. In other words, a classifier using these structural metrics to predict MIA output has high precision but lower recall---although the latter improves if we take into account the statistical significance of MIA observations.
        \end{enumerate}
        This suggests that the structural metrics are proxies for some as yet unquantified latent property of the models, similar to the conditioned loss metric at the heart of LiRA and other state-of-the-art attacks.
    \item We make clear the fundamental difference between being able to rule out release of a model (as being disclosive) and being able to say that it is safe.
\end{enumerate}

These contributions allow us to avoid computationally costly MIA attacks and provide actionable insights into how to reduce disclosure risk. The remainder of this paper is organised as follows. We begin in Section~\ref{sec:background} by discussing MIAs, traditional SDC within TREs, and the implications for ML privacy. Section~\ref{sec:methodology} then details the methodology used to investigate the privacy risks of hyperparameters and structural metrics in tree-based ML models. We present our empirical results in Section~\ref{sec:results}, and finally, Section~\ref{sec:conclusion} summarises and discusses our findings.

%% file: 02_background.tex
\section{Background}%
\label{sec:background}

\subsection{Membership Inference Attacks}

Recognising different sources of uncertainty in the model training process, the MIA can be framed as the question: \textit{what are the probabilities that a model $M$ comes from the population $\mathcal{M}_{\text{in}}$ (or $\mathcal{M}_{\text{out}}$) of models trained with (or without) a particular record $X$?} For this to take a computational form requires a mapping from some set of metrics describing the structure and/or behaviour of models, and records, onto a probability distribution. Given the current state of knowledge, a key issue is how to identify one or more suitable metrics as proxies for underlying risk causes.

A common early approach to MIA exploits the difference in confidence scores between predictions for samples that were in the training set and those that were not~\cite{Shokri:2017}. However, this approach fails to take into account that (hopefully) models will correctly identify regions where high probability records have the same label---whether they are from training set or not. The conceptual and methodological shortfalls of this approach have been well documented~\cite{Rezaie:2021}. 

More recent attacks implicitly focus on decision boundaries by considering the record-wise difficulty of making correct predictions for the original task. For example, LiRA uses as proxy the prediction error (loss) for a record $X$ from the original target model $M$, comparing it to the distributions of loss values from sampled subsets from $\mathcal{M}_{\text{in}}$ and $\mathcal{M}_{\text{out}}$~\cite{Carlini:2022}. Similarly, the quantile regression~\cite{Bertran:2023} and robust MIA (RMIA)~\cite{Zarifzadeh:2024} condition the confidence on that for other records (and models). Relative measures of MIA vulnerability have also been suggested, computing risk based on the difference in prediction loss for member samples relative to a predefined reference model~\cite{Bai:2025}. Recent work has also sought to improve performance with target model knowledge distillation~\cite{Galichin:2025}. 

MIAs typically assess whether a given data point exactly matches a training record, which potentially neglects similar or overlapping memorised data. For example, an image from a different angle may still be recognisable and evoke an exploitable behaviour from the model even though it would be incorrectly assessed as a false positive with a binary exact matching metric. Consequently, Tao and Shokri~\cite{Tao:2025} propose a broader assessment framework that evaluates whether the model was trained on any data within a specified range using domain-specific distance measures.

Typically, MIAs assume that the model's confidence scores (or loss values) are exposed. However, even when this information is hidden or unavailable, label-only based attacks can still identify whether a record was in the original training data. This is achieved by perturbing records to assess model robustness with the assumption that records exhibiting high robustness belong to the training dataset~\cite{Choquette-Choo:2021}. Moreover, in some cases training membership can be accurately inferred with just a single query~\cite{Peng:2024} by exploiting the fact that member samples exhibit more resistance to adversarial perturbations than non-member samples.

One of the most significant drawbacks for assessing ML privacy with MIAs is their computational cost, particularly when using shadow models, which can require training many models with the same architecture and algorithm as the original target model. This can mean that they may not be feasible for large models or datasets. Consequently, a growing body of work is exploring ways to reduce the computational burden, for example maintaining MIA accuracy with only a few shadow models~\cite{Zarifzadeh:2024}. Interestingly, a recent theoretical analysis of likelihood ratio-based MIAs on stochastic gradient descent~\cite{Leemann:2023} has shown how various hyperparameters influence vulnerability, without needing to train any shadow models.

A number of risk factors and proxy measures have been established, with overfitting (quantified by generalisation error) recognised as a significant cause of information leakage~\cite{Yeom:2018, Yeom:2020}. Increasing model complexity often entails a trade-off between privacy and utility, as observed in neural networks where more parameters can enhance performance but compromise privacy~\cite{Tan:2022}. While explainable ML models are often proposed to boost transparency and trust, recent studies caution that they may inadvertently increase vulnerability by revealing feature importance information, which can be exploited~\cite{Liu:2024, Naretto:2025}; perturbing these features typically results in a bigger loss in confidence for member samples.

Implementing robust ML practices to prevent overfitting and ensure generalisation is therefore an important first step in mitigating MIA vulnerabilities~\cite{Dionysiou:2023}. Additionally, a number of defences have been proposed~\cite{Hu:2023, Jarin:2023}. The most common of these is to incorporate information perturbation within the learning algorithm, such as through differential privacy (DP)~\cite{Dwork:2008}. However, the use of DP and knowledge distillation techniques may only mitigate some inference attacks~\cite{Liu:2022}. Indeed it has been suggested that model distillation may not provide any effective privacy defence~\cite{Jagielski:2023a}, and the level of DP necessary to protect a model from MIA may render its utility unusable~\cite{Rahman:2018}.

Moreover, it has recently been highlighted that in practice DP implementations suffer from a number of problems and standard anti-overfitting techniques can often achieve better results~\cite{Blanco-Justicia:2022}. For example, standard regularisation techniques have been shown to help mitigate the risk posed by an increasing number of model parameters~\cite{Tan:2023}.

As the use of ML continues to proliferate, it is becoming increasingly common to release multiple updated versions of a model. However, doing so may further expose confidential properties of the data, strengthening reconstruction attacks~\cite{Salem:2020}. This has also been shown to be true for improving MIA success~\cite{Jagielski:2023b}, which may be a particular risk for secure data facilities such as TREs that may egress many models trained on the same data, and is a similar concern to differencing attacks in traditional output SDC\@.

\subsection{Trusted Research Environments}

Trusted research environments (TREs)\footnote{Also known as safe havens, secure data environments, and secure research services.} offer a means for researchers to analyse confidential datasets and publish their findings. The globally recognised Five Safes\footnote{Safe people, safe projects, safe settings, safe data, safe outputs.} approach to privacy preservation and regulatory compliance~\cite{Green:2023} requires TRE staff to independently check the disclosure risk posed by any proposed output before release is approved.

This process of SDC is well understood for outputs of traditional statistical analyses. However, while the use of ML methods has rapidly expanded, a recent study of TREs by Kavianpour et~al.~\cite{Kavianpour:2022} in the UK revealed a shortfall of SDC understanding and tools to support the use of ML\@. In particular, while researchers can already use ML methods within TREs, the inability to export those trained models precludes reproducibility of results, a critical tenet of scientific practice~\cite{Arnold:2019}.

Regulatory authorities such as the UK Information Commissioner's Office recognise the risks of ML models, but offer general advice rather than practical guidance and tools. Jefferson et~al.~\cite{Jefferson:2022} have recently presented a detailed discussion of the implications of ML model release and proposed a set of guidelines for TREs. A recent report by DARE UK~\cite{Dare:2022} highlighted the need to \textit{``Where possible, automate the review of outputs to support and focus the use of skilled personnel on the areas of most significant risk''}.

\subsection{Relationship to Statistical Disclosure Control}%
\label{sec:structural}

We briefly introduce the notions of SDC and TREs, before turning our attention to the implications of this body of work for the risk assessment of ML models in practice.

\subsubsection{TREs and SDC}
For all the apparent novelty of risk assessing ML models, it is important to place these within a wider context of assessing the risk of privacy leakage from any outputs created from confidential data. From that perspective, risks such as attribute inference date back to at least the instructions for the 1850 US census:

\begin{quote}
    ``to consider the facts intrusted [sic] to them as\ldots~not to be used in any way to the gratification of curiosity, the exposure of any man's business or pursuits\ldots''~\cite{Ruggles:2023}
\end{quote}

Nowadays, national statistics institutes (and other holders of confidential information such as health authorities) routinely conduct SDC of all outputs they produce. They also offer access to researchers via TREs. TREs almost universally employ the Five Safes framework to provide multiple layers of assurance to data owners, within which the final stage of checking for `safe outputs' is a vital part.

Notably, most TREs have long employed principles-based SDC~\cite{Ritchie:2015} with human output checkers involved to facilitate nuance and assess the `discreditability of an attack' in the interpretation of risk analysis. Similar ideas have recently been discussed for ML models~\cite{Rezaei:2023}. 

\subsubsection{Implications for ML Privacy Research}%
\label{subsec:struct_measures}

Thus for traditional forms of analyses, there is decades of well understood theory and practice surrounding the SDC of outputs such as (multi-dimensional) tables, regression models, survival statistics and a range of other outputs, which have recently been consolidated into a coherent `statbarns' framework~\cite{Green:2024}. From these we can identify three key concepts relevant to ML models:

\begin{itemize}
    \item Residual degrees of freedom risk (DoF): for regression models such as linear, logistic or probit, TREs typically require the number of records used to create the model exceeds the number of trainable parameters by at least 10.

        The equivalent of this can be directly calculated for ML models; for example, the number of values involved in specifying branches in (forests of) trees, or the number of trainable weights in neural networks.

    \item $k$-anonymity: queries should not report on groups of fewer than $k$ people. In practice, for tables this means that each cell must reach a minimum threshold count. 

        Since tree-based models (and by extension forests) create a partition of the input space in exactly the same way as tables do, this can be measured directly for a model if we have the training data and the model. For more complex cases we can derive a proxy for this for each training record: the number of other training records for which the model gives the same output (to some precision).

    \item Class disclosure risk. It should not be possible to make inference about all members of a subgroup via the publication of query results containing zeros, e.g., \textit{``in our fictitious survey no ML researchers were aware of the Secure Data Access Professionals manual''}\footnote{In some cases it may be argued that zeros may be structural (people born without certain organs are unlikely to develop cancers of those organs) or may be `evidential' and scientifically important: \textit{``no people with gene XYZ had disease A''}. Hence the practice of applying principles-based SDC rather than strictly applying rules.}.

        For ML models we can interpret class disclosure risks as occurring when the model's predicted probability for \textit{any} class falls below a threshold ($k$/size of training set) for \textit{any} record. In other words, a model that is completely confident in its predictions presents a class disclosure risk. While this might be seen excessively restrictive, the threshold will be small in practice, and we note that this is somewhat similar to the effects produced by the defence strategies proposed by MemGuard~\cite{memguard}. 

        We note in passing that protection against class disclosure is at best poorly catered for by DP training methods since the guarantees ($\epsilon$) multiply with the size of the subgroup, which may not be known in advance.
\end{itemize}

%% file: 03_methodology.tex
\section{Methodology}%
\label{sec:methodology}

This section presents the methodology used to investigate MIA risks in tree-based classification models. We first state our hypotheses for disclosure risk based on model hyperparameters in Section~\ref{sec:method_hyper} and structural model properties in Section~\ref{sec:method_structural}. Section~\ref{sec:tpr_ssd} provides further details of the statistical framework used for determining the significance of observed MIA success. We then describe our experimental setup in Section~\ref{sec:setup}. This includes details of the specific MIA and metrics explored, including a novel metric $\text{FDIF}_z$, as well as target models and datasets used for evaluation.

\subsection{Disclosure Risk and Model Hyperparameters}%
\label{sec:method_hyper}

It is here hypothesised that while absolute disclosure risk is dataset-dependent, the \textit{relative} risk ranking of different model hyperparameter combinations may generalise across datasets. However, since many different risk metrics have been proposed, it is first necessary to (i) select a single representative measure to analyse the relative risk; and (ii) establish that this metric is likely to generalise across different datasets. Subsequently, we test whether a simple, interpretable model can predict if a hyperparameter combination leads to relatively high or low risk as defined by this chosen metric.

That is, we hypothesise:
\begin{itemize}
    \item H1.1: Since MIA metrics measure different aspects of the same phenomenon (risk), hyperparameter rank orderings for one metric will be predictive of rank orderings for another. To test this hypothesis, for each target model algorithm, pair-wise rank correlation statistics are computed to compare the relative risk \textbf{between metrics} across a range of hyperparameter combinations for each dataset.
    \item H1.2: This rank ordering will generalise \textbf{across datasets}. To test this hypothesis, we select a single representative MIA metric and compute Kendall $\tau$ rank correlation coefficients of hyperparameter risk to perform pair-wise comparisons for each dataset.
    \item H1.3: We can exploit H1.1 and H1.2 to produce a \textbf{risk prediction model} for `unnecessary vulnerability' that can be applied to different datasets without the need for sampling MIA scores on these new datasets. To test this hypothesis, we assess the performance of models trained to predict whether a given hyperparameter combination is relatively high or low risk as a binary classification task. These models are trained on a single dataset (mimic2-iaccd) that has been sampled for MIA risk, and the rules are then extracted and evaluated on the remaining (unseen) datasets to test generalisation performance.

        That is, for each target model type, a Decision Tree classifier is used to learn a simple human-interpretable ruleset that predicts whether a hyperparameter combination is relatively high or low risk. Risk is here defined as high if the previously chosen metric is in the top 20\% of all hyperparameter combinations. While this is a somewhat arbitrary threshold, the aim is to prove that a simple set of rules can be identified that predict whether a hyperparameter combination poses unnecessarily high risk.

        In order to extract rules that are easy to interpret, the Decision Trees are restricted to a maximum depth of 5; using a simple grid search for the other hyperparameters. Stratified 10-fold cross-validation was applied to the single training dataset (mimic2-iaccd) and the best performance was found in each case to be a minimum number of samples per split of 2, a minimum number of samples per leaf of 2, the entropy criterion, and a balanced class weight.
\end{itemize}.

\subsection{Disclosure Risk and Structural Model Properties}%
\label{sec:method_structural}

To investigate the relationship of disclosure risk and the structural model properties described in Section~\ref{subsec:struct_measures}, we take the following approach. For a range of algorithm-dataset-hyperparameter combinations sampled, we analyse the model produced and its predictions for the training set, and use these to capture the structural metrics.

Note that different to the \textit{ante-hoc} `unnecessary risk' metric explored above, this is a \textit{post-hoc} analysis, since the minima (e.g., samples per leaf) and maxima (e.g., depth of a tree) defined via an algorithm's hyperparameters will be lower/upper bounds that may not be reached if the data does not merit it.

We then join these with the various metrics captured from the MIAs and use this combined data to examine two complementary hypotheses:

\begin{itemize}
    \item H2.1: Structural risks are \textbf{necessary} conditions for model vulnerability. In other words, if a given structural metric is 0 then there will be no MIA risk. In the following analysis we insist on 100\% support for hypothesis 2.1 because this is a statement about \textbf{safety}.

    \item H2.2: The presence of structural risks are \textbf{sufficient} conditions for model vulnerability. In other words, if (the product of a combination of) structural metric(s) is 1 there will be non-zero MIA risk. We insist on high, but not 100\% support for hypothesis 2.2 because this is a statement about \textbf{risk}.
\end{itemize}   
 
\subsection{Statistical Significance of MIA Vulnerabilities}%
\label{sec:tpr_ssd}

\subsubsection{Preliminaries}

Assume a researcher has access to a dataset $\mathcal{D}$ of size $N=|\mathcal{D}|$ which they split into disjoint training and test sets $\mathcal{D}_{\text{train}}, \mathcal{D}_{\text{test}}$ of sizes $N_{\text{train}}$ and $N_{\text{test}}$ respectively. They use these to train and test a \textit{target} model $M$ for a classification problem with $L$ classes, so $M: Domain(\mathcal{D}) \rightarrow [0,1]^L$. For simplicity we assume the outputs of $M$ have been normalised (e.g., via a softmax function) so that they represent the predicted class probabilities for each record. 

MIAs may be created using synthetic datasets which approximate the underlying distribution from which $\mathcal{D}$ is drawn. It is well-established that the success of MIAs often increases as uncertainties in approximating the data distribution are reduced~\cite{Carlini:2022, Shokri:2017, Ye:2022}. 

Thus, in recognition of the pace at which new MIAs are being invented, in our scenario, we remove those uncertainties to estimate an upper bound on the risk. Hence we explicitly use the sets of outputs (either directly as probabilities or converted to losses) as the basis for conducting different MIAs.

For the purpose of evaluating the statistical significance of attack vulnerabilities, we consider a generic MIA regressor $A$ derived from the target model $M$'s outputs. That is:
\begin{itemize}
    \item Without loss of generality, we assume that an attack creates an MIA regression model $A: M(\mathcal{D}) \rightarrow \mathcal{R}$. Given a target model output $x' = M(x), x \in \mathcal{D}$, we condition the outputs so that $A(x')$: represents the estimated probability that $x \in \mathcal{D}_{\text{train}}$.

    \item These $N$ membership estimates can then be sorted into decreasing order and used to produce a ROC curve, which via interpolation can be probed to produce estimates of the true positive rate (TPR) for different thresholds of false positive rate (FPR)\@.

    \item A specific MIA classifier is created by applying a threshold to the MIA regression model's predictions, above which the label `\textit{member}' is output. We note that the likely accuracy of different MIA classifiers will vary since the difficulty of inferring membership will vary between records, as reported elsewhere~\cite{Carlini:2022, Zarifzadeh:2024}. Therefore the estimation procedure described below will need to be repeated for different thresholds.
\end{itemize}

\subsubsection{From True Positive Rates to Attack Vulnerability}

Following Carlini et~al.~\cite{Carlini:2022} it has become standard practice to compare different MIAs based on values for the TPR at different low FPR regimes. We will denote these TPR@$z$, where typically the FPR $z \in \{0, 0.001, 0.01\}$.

However, from the perspective of an SDC output checker tasked with making a decision on whether to approve the release of the trained model $M$, what would be more helpful is to quantify the likelihood of observing different numerical MIA classifier results (TPR@$z$) by chance.

We assess this via a null hypothesis:

\begin{itemize}
    \item $H_0(z)$: The classifier produced by choosing a threshold $t_z$ such that $\text{FPR}=z$ is making uniform random selections (without replacement) from $\mathcal{D}$ when constructing $\mathcal{X}$ the set of records it predicts to be members: $\mathcal{X}\subset \mathcal{D}\mid \forall  x \in \mathcal{X}, A(x)>t_z$.
\end{itemize} 

To assess the likelihood of this null hypothesis $H_0(z)$ we proceed by first deriving an expression for the number of records $n=|\mathcal{X}|$ the MIA classifier will predict as `members' based on the threshold $t_z$ and on the number of true positives it predicts, TPR@$z$.

Let $\text{TP}_z, \text{FP}_z$ be the number of true/false positive records in $\mathcal{X}$, and assume that when the (floating point) true/false positive rates are interpolated from empirical data we take the floor of those values to map onto specific records included in $\mathcal{X}$. By definition:

\begin{eqnarray}
    n =&\text{TP}_z + \text{FP}_z \nonumber\\
    = & \lfloor(\text{TPR@}z \cdot N_{\text{train}}) + \lfloor (z \cdot N_{\text{test}}) \label{eqn:line-1}
\end{eqnarray}

We then apply a one-tail test for $H_0(z)$, assessing the over-representation of true positives in $\mathcal{X}$, using a hypergeometric distribution to calculate the probability of observing $\text{TP}_z$ or more true positives when using a process that makes selections without replacement from $\mathcal{D}$ and stops as soon as $\text{FP}_z$ incorrect `member' predictions have been made. The null hypothesis $H_0(z)$ is rejected at confidence level $\alpha$ if $\alpha > P(\text{TP}_z^+ | H_0, z)$, where

\begin{eqnarray}
    P(\text{TP}_z^+ | H_0, z) &=& 1 - \sum_{i=0}^{i<\text{TP}_z} P(i|\mathcal{D},n) \\
    &=& 1 - \sum_{i=0}^{i<\text{TP}_z} \frac{ C(N_{\text{train}},i)\; C(N_{\text{test}},\text{FP}_z)}{C(N,(i+\text{FP}_z))} \label{step3} \\
    &=& 1 - \sum_{i=0}^{i<\text{TP}_z} \frac{\frac{N_{\text{train}}!}{i!(N_{\text{train}}-i)!}\frac{N_{\text{test}}!}{\text{FP}_z! (N_{\text{test}}-\text{FP}_z)!}}{\frac{N!}{(i+\text{FP}_z)! (N - i - \text{FP}_z)!}} 
\end{eqnarray}

\noindent where Equation~\ref{step3} recognises that it is $\text{FP}_z$ that is fixed rather than $n$ when applying the hypergeometric distribution, and $C(a,b)= \frac{a!}{b!(a-b)!}$ is the number of ways of selecting $b$ items from a total of $a$ without replacement. 

Thus after some rearrangement we assert that with $(1-\alpha) \times 100$\% confidence, a model $M$ is MIA vulnerable at a FPR of $z$ if:

\begin{equation}
    1 - \alpha < \frac{N_{\text{train}}!N_{\text{test}}!}{N!\text{FP}_z!(N_{\text{test}}-\text{FP}_z)!}\sum_{i=0}^{i<\text{TP}_z} \frac{(i+\text{FP}_z)! (N-i-\text{FP}_z)!}{i!(N_{\text{train}}-i)}\label{vulnerableprob}
\end{equation}\label{eq:vulerability}

\subsection{Experimental Setup}%
\label{sec:setup}
 
\subsubsection{Membership Inference Attack}%
\label{sec:mia_attacks}

Here we use the LiRA~\cite{Carlini:2022} implementation from the open source SACRO-ML framework~\cite{Smith:2024} to perform membership inference and assess the disclosure risk presented by a range of fitted tree-based classification models. LiRA is a fine-grained black-box evaluation metric-based MIA~\cite{Niu:2025} and (especially when using sufficient shadow models) is representative of the current state-of-the-art. Furthermore, it places an emphasis on an attacker scenario that corresponds to TRE concerns: \textit{``If a membership inference attack can reliably violate the privacy of even just a few users in a sensitive dataset, it has succeeded''}. For each target model hyperparameter combination, 10 random training and test data splits are performed to fit 10 different models, and a LiRA attack is executed for each. The attack uses 100 shadow models to estimate the loss distributions.   

\subsubsection{Disclosure Metrics}%
\label{sec:mia_metrics}

Since the ML privacy community has yet to reach a consensus on a single disclosure risk metric, a range of standard MIA success metrics are computed from the LiRA outputs; for example, TPR at low FPR, and area under the curve (AUC). These metrics are averaged over the 10 runs to produce a set of risk metrics for a given target model hyperparameter combination on a dataset.

Moreover, we note that attackers will almost always have access to some records they know are non-members of the training set---either because they have artificially created them, or perhaps because they may be their own records. We therefore posit that armed with that information, and aware (from the literature) that in most cases the MIA cannot make a confident prediction, attackers will prefer a model that: (i) does not make high confidence member predictions for records they know are non-members; (ii) is correct when it predicts non-member with high confidence. 

From these premises we derive a novel frequency difference metric $\text{FDIF}_z$, calculated as follows:
    \begin{enumerate}
        \item For each training record we note: the probability (softmax value) of the MIA's prediction for the class member; and the true (binary) value.
        \item This list is sorted by decreasing confidence of the member prediction.
        \item The difference between the proportion of true positives in the extreme subsets; that is, the top and bottom $z \times N$ records respectively is calculated. The more positive this $\text{FDIF}_z$ value, the better the MIA is at discriminating when it is confident. Here we report scores for $z \in \{0.02,0.01,0.001\}$.
        \item Finally we calculate $P(\text{FDIF}_z)$: the probability of observing this (or greater) value as an effect of stochastic sampling if the MIA was making purely random guesses at the extrema.
    \end{enumerate}

In addition to these MIA-specific metrics, several target model metrics (e.g., generalisation error) are also examined for correlations. All metrics considered here are described in Table~\ref{table:metrics}.

\input{table-1.tex}

\subsubsection{Target Models}

The target models assessed here are Random Forest and Decision Tree classifiers from the widely used open source Python scikit-learn~\cite{Pedregosa:2011} package, and the XGBoost classifier~\cite{Chen:2016}. The hyperparameter space is sampled with a range of values for the most commonly used hyperparameters; all others are set to their implementation defaults.

The sets of values used to define all the possible combinations of hyperparameter settings for the Decision Tree classifier, Random forest classifier and XGBoost are given in Table~\ref{table:parameters_combined}.

\input{table-2.tex}

Tree-based models have been shown to provide state-of-the-art performance for tabular data classification tasks~\cite{Schwartz-Ziv:2022, Chen:2016} and therefore a selection of tabular datasets has been chosen here to provide a range of different landscape complexities for target model fitting. Binary classification problems are used here to aid comparison; increasing class number is known to increase the vulnerability to MIA~\cite{Shokri:2017}. Each dataset is split into three partitions, stratified by target label. The first subset is used for target model fitting, and the second is used for assessing target model accuracy. The LiRA attack then uses a combination of both the known training set and the third (unseen) subset to evaluate the MIA success.
 
\subsubsection{Datasets}

The mimic2-iaccd dataset was created for the purpose of investigating the effectiveness of indwelling arterial catheters in hemodynamically stable patients with respiratory failure for mortality outcomes. The dataset is derived from MIMIC-II~\cite{Raffa:2016}, the publicly accessible critical care database. It contains summary clinical data and outcomes for 1776 patients. The following repetitive and uninformative features were removed: `service unit', `day icu intime', `hosp exp flg', `icu exp flg', `day 28 flg' and `sepsis flg'. The target label selected was `censor flg'.

The in-hospital-mortality dataset was created to characterise the predictors of in-hospital mortality for intensive care unit admitted patients. Data were extracted from the MIMIC-III~\cite{Zhou:2021} database from 1177 heart failure patients. Features `ID' and `group' were removed and the target label selected was `outcome'. 
 
The Indian liver dataset~\cite{Ramana:2022}, contains 416 liver patient records and 167 non liver patient records. The dataset was collected from north east of Andhra Pradesh, India. This dataset contains 441 male patient records and 142 female patient records. The feature `gender' was transformed to numerical values and the target label selected was ‘Selector’, which identifies whether patients have liver disease or not. 
 
The synth-ae dataset~\cite{SyntheticData} was created by a Bayesian statistical model of de-identified patient information collected in the English National Health Service from 2014 to 2018. The following features were removed: `AE Arrivea Date', `AE Arrive HourOfDay', `Admission Method', `ICD10 Chapter Code', `Treatment Function Code', `Length Of Stay Days', `ProvID'. Any patients with missing data were also removed. Non-numerical features were transformed to numerical. Only the first 5000 rows were kept to generate the models. The target label selected was `Admitted flag'.

The sick and mammography datasets are open datasets available from OpenML\footnote{\url{https://openml.org}} with ID 41946 and 310, respectively. 

%% file: figs/table-1.tex
\begin{table}[t]
    \caption{MIA and target model metrics.}%
    \label{table:metrics}
    \small
    \centering
    \begin{tabularx}{\textwidth}{lX}
        \toprule
        {\bf Metric} & {\bf Description} \\
        \midrule
        MIA AUC & Area under the receiver operating characteristic curve (AUC) of MIA success. \\
        MIA $P$(AUC) & Probability of observing MIA AUC with a random attack.\\
        MIA $\text{FDIF}_z$ & Difference between MIA TPR in the top and bottom $z$ records. \\
        MIA $P(\text{FDIF}_z)$ & Probability of observing $\text{FDIF}_z$ with a random attack.\\
        MIA $\text{TPR@}z$ & MIA TPR at a FPR of $z: z\in\{0.1, 0.001\}$\%. \\
        MIA Advantage & Maximum FPR minus TPR~\cite{Yeom:2018}. \\
        Target Train ACC & Target model training accuracy. \\
        Target Test ACC & Target model test accuracy. \\
        Target AUC & Target model test AUC\@. \\
        Target GE & Target model generalisation error (GE). While overfitting has been shown to be a sufficient condition for MIA vulnerability in popular ML algorithms, it is not a necessary condition~\cite{Yeom:2018}. GE can therefore be considered as a lower bound for disclosure risk. \\
        \bottomrule
    \end{tabularx}
\end{table}

%% file: figs/table-2.tex
\begin{table}[t]
    \caption{Target models and hyperparameters sampled.}%
    \label{table:parameters_combined}
    \small
    \centering
    \begin{tabular}{lll}
        \toprule
        {\bf Model} & {\bf Name} & {\bf Values}\\
        \midrule
        \multirow{7}{*}{Decision Tree} 
        & \texttt{criterion} & gini, entropy\\
        & \texttt{max\_depth} & 1, 2, 5, 10, 20\\
        & \texttt{min\_samples\_leaf} & 1, 2, 5, 10, 20\\
        & \texttt{min\_samples\_split} & 2, 5, 10, 20\\
        & \texttt{max\_features} & None, sqrt, log2\\
        & \texttt{class\_weight} & None, balanced\\
        & \texttt{splitter} & best, random\\
        \midrule
        \multirow{8}{*}{Random Forest} 
        & \texttt{criterion} & gini, entropy\\
        & \texttt{max\_depth} & 1, 2, 5, 10, 20\\
        & \texttt{n\_estimators} & 2, 5, 10, 20, 50, 100\\
        & \texttt{min\_samples\_split} & 2, 5, 10, 20\\
        & \texttt{class\_weight} & None, balanced\\
        & \texttt{min\_samples\_leaf} & 1, 2, 5, 10, 20\\
        & \texttt{bootstrap} & True, False\\
        & \texttt{max\_features} & None, sqrt, log2\\
        \midrule
        \multirow{4}{*}{XGBoost} 
        & \texttt{n\_estimators} & 1, 2, 5, 10, 15, 25, 50, 75, 90, 100\\
        & \texttt{max\_depth} & 1, 2, 5, 10, 20\\
        & \texttt{reg\_alpha} & 0, 0.001, 0.005, 0.01, 0.05\\
        & \texttt{min\_child\_weight} & 1, 2, 4, 8, 12\\
        \bottomrule
    \end{tabular}
\end{table}

%% file: 04_results.tex
\section{Results}%
\label{sec:results}

\subsection{Results for Model Hyperparameter Risks}%

\subsubsection{Risk Correlation Across Metrics}

Table~\ref{table:kendall_mimic} shows the Kendall $\tau$ rank correlation statistics for each metric with pair-wise comparisons for the XGBoost classifier averaged over all datasets. Strong correlations between all of the MIA metrics can be seen. Target model generalisation error also positively correlates with risk, however target model test accuracy does not, indicating that it is strongly dataset dependant. These correlations are also seen on the other datasets with both Random Forest and Decision Tree classifiers (not shown). Since these metrics are highly correlated, this provides strong evidence in support of hypothesis~1.1 and the MIA AUC is used for simplicity in the remainder of the analysis. Regardless of the chosen metric, the following results should hold.

\input{table-3.tex}

\subsubsection{Risk Correlation Across Datasets}

Table~\ref{table:kendall_combined} shows the correlation of risk across different datasets to test hypothesis~1.2. That is, the Kendall $\tau$ rank correlation statistics for dataset pair-wise comparisons of the MIA AUC for each target model hyperparameter combination. As can be seen, there is a strong correlation of risk with respect to hyperparameter combination across the different datasets for each of the target model types. For Decision Tree models the mean is 0.64 (min=0.45, max=0.81); for Random Forest models the mean is 0.72 (min=0.51, max=0.88); and for XGBoost models the mean is 0.67 (min=0.48, max=0.84). This suggests that it may be possible to train a classification model that identifies the relative risk of a hyperparameter combination for most datasets.

\input{table-4.tex}

\subsubsection{Risk Prediction Across Datasets}

Table~\ref{table:accuracy_combined} shows the precision, recall, and accuracy of the simple rulesets at predicting whether a given target model hyperparameter combination is in the top 20\% highest MIA AUC (high risk) or in the bottom 80\% (low risk). Note that the risk prediction rules for each target model were derived by training a Decision Tree classifier on the results from the mimic2-iaccd dataset then distilling the tree into a set of rules. The mean accuracy on the unseen datasets is 91.4\% (min=89.1\%, max=93.9\%) for Decision Tree target models; 95.2\% (min=92.9\%, max=96.2\%) for Random Forest; and 94.1\% (min=90.8\%, max=97.2\%) for XGBoost target models, showing that the very simple ruleset is able to accurately differentiate the hyperparameter combinations with the highest relative risk on each dataset, supporting hypothesis~1.3.

\input{table-5.tex}

Scatter plots showing the relationship between target model accuracy (test set AUC) and privacy risk (MIA AUC) for different hyperparameter combinations of the Decision Tree classifier can be seen in Figure~\ref{fig:scatter_dt}. Similar plots are shown for the Random Forest classifier target model in Figure~\ref{fig:scatter_rf} and for the XGBoost classifier target model in Figure~\ref{fig:scatter_xgboost}. The high accuracy of the risk rulesets is illustrated by colouring the points red where the rules predict a hyperparameter combination to have an MIA AUC in the top 20\% highest risk. Overall, it can clearly be seen that removing the high risk combinations would eliminate a large portion of significantly vulnerable models.

Interestingly, the plots show that removing the hyperparameter combinations predicted as high risk results in little to no loss of maximum achievable target AUC\@ for most datasets. In the few cases where there is a hyperparameter combination classified as high risk that results in a higher target AUC, the difference is small, yet the reduction in MIA AUC is much larger. For example, the Decision Tree target model hyperparameter combinations with the highest accuracy are categorised as low risk for every dataset except the hospital dataset. Moreover, on the hospital dataset, the loss of accuracy (AUC) that would be experienced by removing the high risk combinations is only 0.0011. As another example, removing the high risk hyperparameters for Random Forest target models on the sick dataset, would result in a reduction in the maximum achievable accuracy (AUC) of only 0.005, but a reduction in MIA AUC of 0.13. 

\input{figure-1.tex}
\input{figure-2.tex}
\input{figure-3.tex}

Beyond overfitting, MIA vulnerability is significantly influenced by model type and structure~\cite{Shokri:2017}. This is confirmed here with the rules learned from the mimic dataset for identifying high risk hyperparameter combinations in tree-based models. As might be expected, these rules consistently associate `high risk' MIA vulnerability with model configurations that promote higher complexity. For example, across all model types (Decision Tree, Random Forest, and XGBoost), a greater \texttt{max\_depth} is a recurring condition for high risk. Similarly, a larger number of \texttt{n\_estimators} in ensemble methods (Random Forest and XGBoost) also points to increased vulnerability.

However, the rules reveal nuances beyond sheer size. Specific hyperparameter choices that dictate the model's learning capacity and flexibility are critical:

\begin{itemize}
    \item For Decision Trees, risk factors include: utilising \texttt{splitter=best} rather than \texttt{random}; allowing small values for \texttt{min\_samples\_leaf} or \texttt{min\_samples\_split}; or considering all features for splits (\texttt{max\_features=None}).
    \item In Random Forests, conditions such as \texttt{bootstrap=False} (where each tree is trained on the entire dataset, potentially reducing diversity) or specific \texttt{min\_samples\_split} thresholds when combined with other hyperparameters contribute to high risk.
    \item For XGBoost, a low \texttt{min\_child\_weight}, which permits the creation of tree nodes with fewer samples, is a key indicator of increased vulnerability.
\end{itemize}
 
These specific hyperparameter settings effectively enable the models to create more intricate and numerous decision boundaries. This increased capacity to fit the training data closely not only reflects a higher Vapnik-Chervonenkis (VC) dimension (hence the link to generalisation error) but also, as the rules suggest, heightens the model's susceptibility to MIA\@.

The greediness of the search algorithm also significantly contributes to risk with \texttt{splitter} and \texttt{bootstrap} controlling how the trees are built. The \texttt{splitter} parameter determines the strategy used to choose the best split point for each node in a tree and a greedy approach evaluates all possible splits and chooses the one that maximises the chosen criterion. This can lead to trees that are highly sensitive to the specific training data and may not generalise well. Similarly, \texttt{bootstrap} controls whether the individual trees are trained on random subsets of the training data, which can introduce diversity into the ensemble, reducing overfitting. If bootstrapping is used in a way that reduces diversity, the trees may become too similar and prone to the same overfitting tendencies, again increasing risk.

A combination of higher \textit{potential} model complexity and greedy search algorithms are therefore much more likely to lead to models that are highly susceptible to overfitting and thus pose a higher disclosure risk in practice. The risk prediction rules may be found in Appendix~\ref{sec:rules}.

\subsection{Results for Structural Risks}%
\label{sec:structural-results}

To examine support for the two different hypotheses stated in Section~\ref{sec:method_structural} we aggregated the results from LiRA and structural attacks on all (73610) combinations of dataset-algorithm-hyperparameters. We also created a \texttt{Combined\_AND} structural risk as the logical conjunction (AND) of the degrees of freedom (\texttt{DoF}), $k$-anonymity (\texttt{K\_anon}) and class disclosure (\texttt{lowvals\_cd}) structural measures.

Figure~\ref{fig:violin-dataset-metric} shows the histograms of the observed values of different MIA metrics (columns) for different datasets (rows) separated into models with combined structural risk 0 (low risk; green bars) and 1 (high risk; red bars). 

\input{figure-4.tex}

As can be seen:
\begin{itemize}
    \item The risk metrics have different distributions; higher where there is an identified combined structural risk (red).
    \item Except for the TPR@0.001 metric and the two biggest datasets, there are always cases with no combined structural risk but non-zero MIA risk (green bars).
    \item In contrast, only in a tiny number of cases is there evidence of models having 0 value for MIA risk metrics but 1 for the combined structural risk metric (red bars).
\end{itemize}

Hypothesis 2.1 proposed that when there was no structural risk, the LiRA risk metrics would also be near zero. The second observation above is illustrated in Figure~\ref{fig:heatmaps} (left). This shows conclusive evidence that this hypothesis can be rejected. In other words, \textbf{structural risk is not necessary} for vulnerability to MIA, at least in the form in which structural risk is currently quantified.

Hypothesis 2.2 took the alternative position that the presence of structural risk was predictive of a model being vulnerable to LiRA. As Figure~\ref{fig:heatmaps} (right) shows there are only a handful of such cases that do not also display MIA vulnerability. 

\input{figure-5.tex}

While these counterexamples cannot be ignored; from the perspective of an analyst making pragmatic decisions about whether a trained model poses a privacy risk, there is a clear message. The combined structural measure, \texttt{Combined\_AND} presents \textbf{sufficient but not necessary} evidence of MIA vulnerability, without the need to run computationally expensive attacks. A risk prediction classifier based on this simple rule has:
\begin{itemize}
    \item high precision (0.927--0.993, depending on the specific MIA risk metric being predicted)
    \item but relatively low recall (0.461--0.514, depending on the attack metric).
\end{itemize} 
 It is worth reiterating that this is only a filtering approach. A prediction of `not unsafe' should not be interpreted as meaning the target is safe, rather that further investigation is needed (because the \texttt{AND} rule has low recall). 
This is somewhat akin to the distinction between `innocent', `guilty', and `not proven' in some legal jurisdictions.

\subsubsection{Analysis of Where MIA Metrics are Statistically Significant}

Section~\ref{sec:tpr_ssd} describes a procedure for calculating whether the TPR at a given FPR is in fact statistically significant. For the FDIF metric we also have the equivalent calculation of probability that the result is observed by chance. Figure~\ref{fig:ecdf_metrics_ssd} shows the results (95\% confidence level) of using this analysis to mask the results (i.e., metric values are set to 0 if the null hypothesis cannot be rejected) and then plotting the cumulative distribution of observed metric values. From these plots we can summarise that:

\input{figure-6.tex}

\begin{itemize}
    \item In the overwhelming majority of cases, the numbers of true positive cases identified by LiRA for a fixed (low) proportion of false positives are not high enough to reject the null hypothesis that the attack is just making random guesses. Of the 73600 different algorithm-dataset-hyperparameter combinations tested, only 451 were statistically significant for TPR@0.01 and 97 for TPR@0.001. \\
In Figure~\ref{fig:ecdf_metrics_ssd} this is evidenced in the two left hand plots. The distributions of observed $\text{TPR@}z$ metrics for raw data (ignoring statistical significance) show a clear difference between whether structural risk is present (plain red line) or not (plain green). However that difference disappears for the significant results (lines with markers).
    \item By way of contrast, the FDIF metrics are almost always significant, i.e., the pairs of plain and cross-marked lines are close. The main difference is for $\text{FDIF}_{0.001}$ when only the top and bottom 0.1\% of records (ranked by attack confidence) are considered. Even in this case the difference between the extrema is significant in more than 55\% of cases where the structural metric is 0 and 90\% of cases where the structural risk metric is 1.

    \item For all cases, the cumulative frequency of cases where the structural metric is 1 (red line) grows more slowly than when it is 0. In other words, LiRA risk metrics are higher when the structural risk is present.
\end{itemize}

%% file: figs/table-3.tex
\begin{table}[t]
    \caption{Kendall $\tau$ rank correlation statistics of different MIA and target model metrics for XGBoost target models. Mean over all datasets. Darker shades of grey indicate stronger correlation.}%
    \label{table:kendall_mimic}
    \scriptsize
    \centering
    \begin{tabular}{lrrrrrrr|rrrr}
        & \multicolumn{7}{c|}{{\bf MIA}} & \multicolumn{4}{c}{{\bf Target Accuracy}} \\
        & AUC & $P(\text{AUC})$ & $P(\text{FDIF}_{01})$ & FDIF$_{01}$ & \multicolumn{2}{c}{\stackanchor{TPR}{\hspace{1mm} @0.1 \hspace{2mm} @0.001}} & Adv. & Train & Test & AUC & GE \\
        \midrule
         MIA AUC & \cellcolorgrey{} & \cellcolorgrey{-0.90} & \cellcolorgrey{0.86} & \cellcolorgrey{0.87} & \cellcolorgrey{0.86} & \cellcolorgrey{0.72} & \cellcolorgrey{0.90} & \cellcolorgrey{0.83} & \cellcolorgrey{0.27} & \cellcolorgrey{0.30} & \cellcolorgrey{0.80} \\
         MIA $P(\text{AUC})$ & \cellcolorgrey{-0.90} & \cellcolorgrey{} & \cellcolorgrey{-0.81} & \cellcolorgrey{-0.82} & \cellcolorgrey{-0.82} & \cellcolorgrey{-0.67} & \cellcolorgrey{-0.85} & \cellcolorgrey{-0.79} & \cellcolorgrey{-0.28} & \cellcolorgrey{-0.31} & \cellcolorgrey{-0.76} \\
         MIA $P(\text{FDIF}_{01})$ & \cellcolorgrey{0.86} & \cellcolorgrey{-0.81} & \cellcolorgrey{} & \cellcolorgrey{0.96} & \cellcolorgrey{0.86} & \cellcolorgrey{0.74} & \cellcolorgrey{0.82} & \cellcolorgrey{0.83} & \cellcolorgrey{0.28} & \cellcolorgrey{0.26} & \cellcolorgrey{0.80} \\
         MIA FDIF$_{01}$ & \cellcolorgrey{0.87} & \cellcolorgrey{-0.82} & \cellcolorgrey{0.96} & \cellcolorgrey{} & \cellcolorgrey{0.88} & \cellcolorgrey{0.75} & \cellcolorgrey{0.84} & \cellcolorgrey{0.83} & \cellcolorgrey{0.28} & \cellcolorgrey{0.26} & \cellcolorgrey{0.80} \\
         MIA TPR@0.1 & \cellcolorgrey{0.86} & \cellcolorgrey{-0.82} & \cellcolorgrey{0.86} & \cellcolorgrey{0.88} & \cellcolorgrey{} & \cellcolorgrey{0.73} & \cellcolorgrey{0.83} & \cellcolorgrey{0.81} & \cellcolorgrey{0.27} & \cellcolorgrey{0.26} & \cellcolorgrey{0.78} \\
         MIA TPR@0.001 & \cellcolorgrey{0.72} & \cellcolorgrey{-0.67} & \cellcolorgrey{0.74} & \cellcolorgrey{0.75} & \cellcolorgrey{0.73} & \cellcolorgrey{} & \cellcolorgrey{0.71} & \cellcolorgrey{0.73} & \cellcolorgrey{0.28} & \cellcolorgrey{0.20} & \cellcolorgrey{0.70} \\
         MIA Advantage & \cellcolorgrey{0.90} & \cellcolorgrey{-0.85} & \cellcolorgrey{0.82} & \cellcolorgrey{0.84} & \cellcolorgrey{0.83} & \cellcolorgrey{0.71} & \cellcolorgrey{} & \cellcolorgrey{0.82} & \cellcolorgrey{0.28} & \cellcolorgrey{0.30} & \cellcolorgrey{0.78} \\
         Target Train & \cellcolorgrey{0.83} & \cellcolorgrey{-0.79} & \cellcolorgrey{0.83} & \cellcolorgrey{0.83} & \cellcolorgrey{0.81} & \cellcolorgrey{0.73} & \cellcolorgrey{0.82} & \cellcolorgrey{} & \cellcolorgrey{0.32} & \cellcolorgrey{0.27} & \cellcolorgrey{0.90} \\
         Target Test & \cellcolorgrey{0.27} & \cellcolorgrey{-0.28} & \cellcolorgrey{0.28} & \cellcolorgrey{0.28} & \cellcolorgrey{0.27} & \cellcolorgrey{0.28} & \cellcolorgrey{0.28} & \cellcolorgrey{0.32} & \cellcolorgrey{} & \cellcolorgrey{0.61} & \cellcolorgrey{0.22} \\
         Target AUC & \cellcolorgrey{0.30} & \cellcolorgrey{-0.31} & \cellcolorgrey{0.26} & \cellcolorgrey{0.26} & \cellcolorgrey{0.26} & \cellcolorgrey{0.20} & \cellcolorgrey{0.30} & \cellcolorgrey{0.27} & \cellcolorgrey{0.61} & \cellcolorgrey{} & \cellcolorgrey{0.22} \\
         Target GE & \cellcolorgrey{0.80} & \cellcolorgrey{-0.76} & \cellcolorgrey{0.80} & \cellcolorgrey{0.80} & \cellcolorgrey{0.78} & \cellcolorgrey{0.70} & \cellcolorgrey{0.78} & \cellcolorgrey{0.90} & \cellcolorgrey{0.22} & \cellcolorgrey{0.22} & \cellcolorgrey{} \\
        \bottomrule
    \end{tabular}
\end{table}

%% file: figs/table-4.tex
\begin{table}[t!]
    \caption{Kendall $\tau$ rank correlation statistics for MIA AUC across different datasets for each target model type. Darker shades of grey indicate stronger correlation.}%
    \label{table:kendall_combined}
    \small
    \centering
    \begin{tabular}{llrrrrrr}
        \toprule
        {\bf Model} & {\bf Dataset} & {\bf Mimic} & {\bf Liver} & {\bf Hospital} & {\bf Synth} & {\bf Sick} & {\bf Mammo.}\\
        \midrule
        \multirow{6}{*}{Decision Tree} 
        & Mimic &  & \cellcolorgrey{0.81} & \cellcolorgrey{0.76} & \cellcolorgrey{0.79} & \cellcolorgrey{0.56} & \cellcolorgrey{0.62}\\
        & Liver & \cellcolorgrey{0.81} & & \cellcolorgrey{0.73} & \cellcolorgrey{0.73} & \cellcolorgrey{0.53} & \cellcolorgrey{0.60}\\
        & Hospital & \cellcolorgrey{0.76} & \cellcolorgrey{0.73} & & \cellcolorgrey{0.68} & \cellcolorgrey{0.51} & \cellcolorgrey{0.55}\\
        & Synth & \cellcolorgrey{0.79} & \cellcolorgrey{0.73} & \cellcolorgrey{0.68} & & \cellcolorgrey{0.56} & \cellcolorgrey{0.68}\\
        & Sick & \cellcolorgrey{0.56} & \cellcolorgrey{0.53} & \cellcolorgrey{0.51} & \cellcolorgrey{0.56} & & \cellcolorgrey{0.45}\\
        & Mammography & \cellcolorgrey{0.62} & \cellcolorgrey{0.60} & \cellcolorgrey{0.55} & \cellcolorgrey{0.68} & \cellcolorgrey{0.45} &\\
        \midrule
        \multirow{6}{*}{Random Forest} 
        & Mimic & & \cellcolorgrey{0.88} & \cellcolorgrey{0.71} & \cellcolorgrey{0.73} & \cellcolorgrey{0.80} & \cellcolorgrey{0.76}\\
        & Liver & \cellcolorgrey{0.88} & & \cellcolorgrey{0.71} & \cellcolorgrey{0.74} & \cellcolorgrey{0.78} & \cellcolorgrey{0.76}\\
        & Hospital & \cellcolorgrey{0.71} & \cellcolorgrey{0.71} & & \cellcolorgrey{0.51} & \cellcolorgrey{0.67} & \cellcolorgrey{0.61}\\
        & Synth & \cellcolorgrey{0.73} & \cellcolorgrey{0.74} & \cellcolorgrey{0.51} & & \cellcolorgrey{0.68} & \cellcolorgrey{0.73}\\
        & Sick & \cellcolorgrey{0.80} & \cellcolorgrey{0.78} & \cellcolorgrey{0.67} & \cellcolorgrey{0.68} & & \cellcolorgrey{0.71}\\
        & Mammography & \cellcolorgrey{0.76} & \cellcolorgrey{0.76} & \cellcolorgrey{0.61} & \cellcolorgrey{0.73} & \cellcolorgrey{0.71} & \\ 
        \midrule
        \multirow{6}{*}{XGBoost} 
        & Mimic & & \cellcolorgrey{0.84} & \cellcolorgrey{0.61} & \cellcolorgrey{0.79} & \cellcolorgrey{0.75} & \cellcolorgrey{0.76}\\
        & Liver & \cellcolorgrey{0.84} & & \cellcolorgrey{0.74} & \cellcolorgrey{0.68} & \cellcolorgrey{0.72} & \cellcolorgrey{0.69}\\
        & Hospital & \cellcolorgrey{0.61} & \cellcolorgrey{0.74} & & \cellcolorgrey{0.48} & \cellcolorgrey{0.53} & \cellcolorgrey{0.52}\\
        & Synth & \cellcolorgrey{0.79} & \cellcolorgrey{0.68} & \cellcolorgrey{0.48} & & \cellcolorgrey{0.61} & \cellcolorgrey{0.65}\\
        & Sick & \cellcolorgrey{0.75} & \cellcolorgrey{0.72} & \cellcolorgrey{0.53} & \cellcolorgrey{0.61} & & \cellcolorgrey{0.73}\\
        & Mammography & \cellcolorgrey{0.76} & \cellcolorgrey{0.69} & \cellcolorgrey{0.52} & \cellcolorgrey{0.65} & \cellcolorgrey{0.73} &\\ 
        \bottomrule
    \end{tabular}
\end{table}

%% file: figs/table-5.tex
\begin{table}[t!]
    \caption{Accuracy of MIA (high/low) risk prediction rules extracted from the mimic dataset and applied to other datasets for each target model type.}%
    \label{table:accuracy_combined}
    \small
    \centering
    \begin{tabular}{llrrrrr}
        \toprule
        \multirow{2}{*}{{\bf Model}} & \multirow{2}{*}{{\bf Dataset}} & \multicolumn{2}{c}{{\bf Low Risk}} & \multicolumn{2}{c}{{\bf High Risk}} & \multirow{2}{*}{{\bf Accuracy}} \\
        & & {\bf Precision} & {\bf Recall} & {\bf Precision} & {\bf Recall} & \\
        \midrule
        \multirow{6}{*}{Decision Tree} 
        & Mimic (train) & 1.00 & 0.93 & 0.79 & 1.00 & 0.95\\
        & Liver & 0.99 & 0.92 & 0.75 & 0.95 & 0.93\\
        & Hospital & 0.96 & 0.90 & 0.68 & 0.86 & 0.89\\
        & Synth & 0.99 & 0.93 & 0.77 & 0.98 & 0.94\\
        & Sick & 0.97 & 0.91 & 0.71 & 0.90 & 0.91\\
        & Mammography & 0.97 & 0.91 & 0.71 & 0.90 & 0.91\\
        \midrule
        \multirow{6}{*}{Random Forest} 
        & Mimic (train) & 1.00 & 0.98 & 0.91 & 0.99 & 0.98\\
        & Liver & 0.99 & 0.96 & 0.87 & 0.94 & 0.96\\
        & Hospital & 0.98 & 0.96 & 0.86 & 0.93 & 0.96\\
        & Synth & 0.97 & 0.94 & 0.80 & 0.87 & 0.93\\
        & Sick & 0.99 & 0.97 & 0.87 & 0.95 & 0.96\\
        & Mammography & 0.98 & 0.96 & 0.86 & 0.94 & 0.96\\
        \midrule
        \multirow{6}{*}{XGBoost} 
        & Mimic (train) & 1.00 & 0.99 & 0.97 & 0.99 & 0.99\\
        & Liver & 0.98 & 0.98 & 0.92 & 0.94 & 0.97\\
        & Hospital & 0.96 & 0.96 & 0.82 & 0.84 & 0.93\\
        & Synth & 0.97 & 0.97 & 0.86 & 0.88 & 0.95\\
        & Sick & 0.94 & 0.94 & 0.76 & 0.78 & 0.91\\
        & Mammography & 0.97 & 0.96 & 0.85 & 0.87 & 0.94\\
        \bottomrule
    \end{tabular}
\end{table}

%% file: figs/figure-1.tex
\begin{figure}
    \subfloat[mimi2-iaccd]{\includegraphics[width=0.48\linewidth]{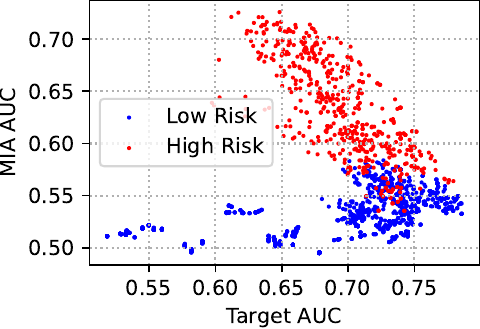}}
    \hfill
    \subfloat[hospital]{\includegraphics[width=0.48\linewidth]{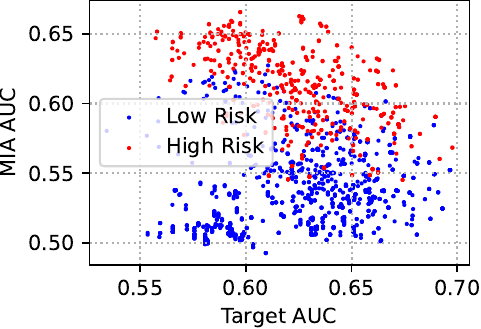}}\\
    \subfloat[indian liver]{\includegraphics[width=0.48\linewidth]{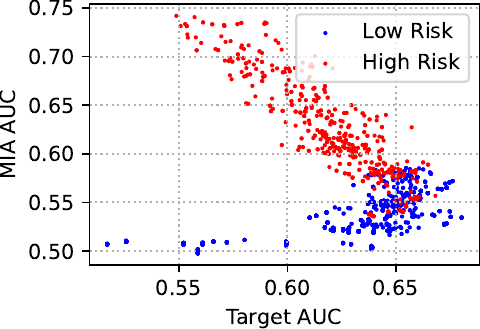}}
    \hfill
    \subfloat[mammography]{\includegraphics[width=0.48\linewidth]{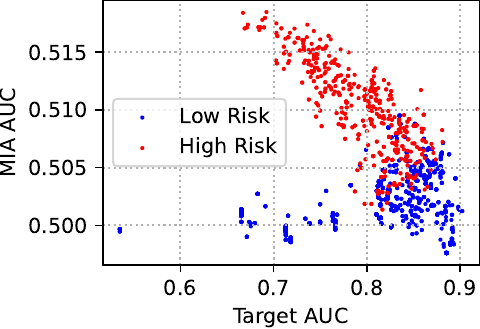}}\\
    \subfloat[sick]{\includegraphics[width=0.48\linewidth]{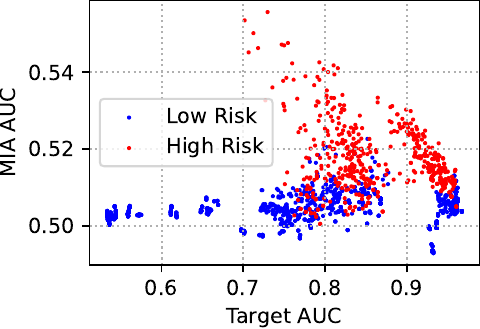}}
    \hfill
    \subfloat[synth-ae]{\includegraphics[width=0.48\linewidth]{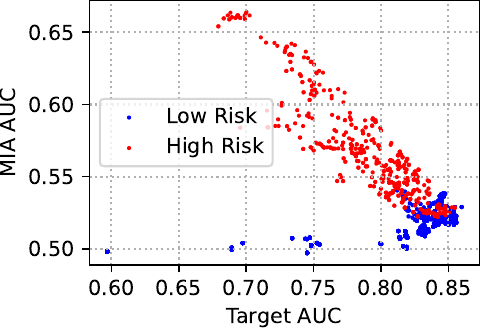}}
    \caption{Privacy risk (MIA AUC) vs.\ accuracy (target AUC) for Decision Tree target models. Points in red denote hyperparameter combinations predicted to be in the top 20\% (high risk) most vulnerable to MIA.}%
    \label{fig:scatter_dt}
\end{figure}

%% file: figs/figure-2.tex
\begin{figure}
    \subfloat[mimi2-iaccd]{\includegraphics[width=0.48\linewidth]{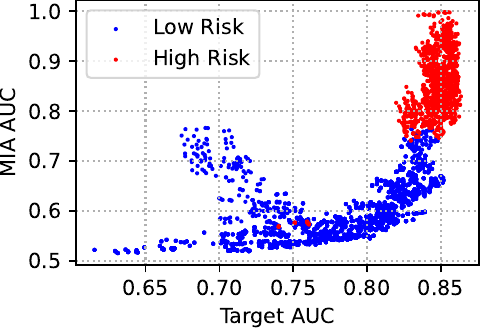}}
    \hfill
    \subfloat[hospital]{\includegraphics[width=0.48\linewidth]{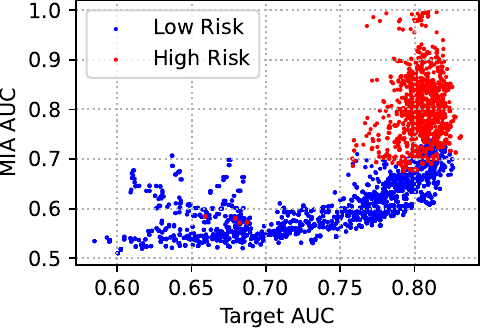}}\\
    \subfloat[indian liver]{\includegraphics[width=0.48\linewidth]{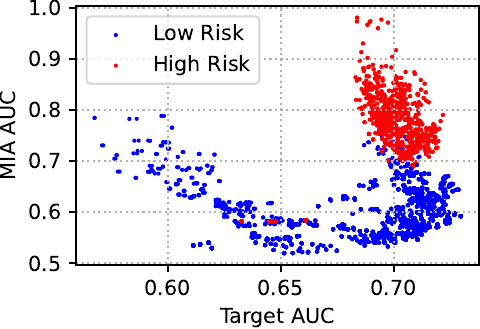}}
    \hfill
    \subfloat[mammography]{\includegraphics[width=0.48\linewidth]{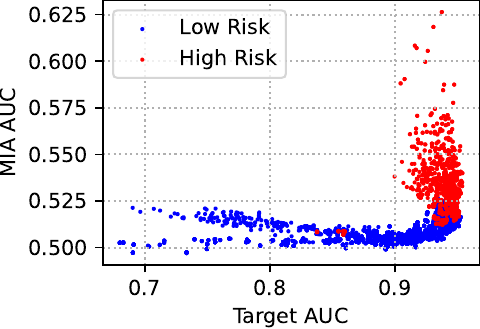}}\\
    \subfloat[sick]{\includegraphics[width=0.48\linewidth]{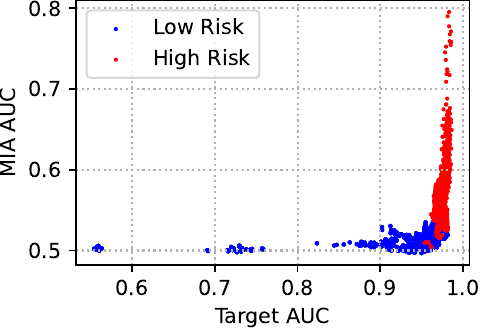}}
    \hfill
    \subfloat[synth-ae]{\includegraphics[width=0.48\linewidth]{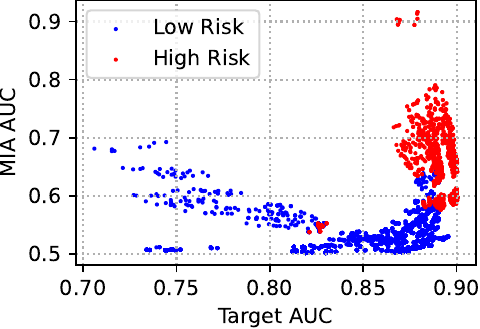}}
    \caption{Privacy risk (MIA AUC) vs.\ accuracy (target AUC) for Random Forest target models. Points in red denote hyperparameter combinations predicted to be in the top 20\% (high risk) most vulnerable to MIA.}%
    \label{fig:scatter_rf}
\end{figure}

%% file: figs/figure-3.tex
\begin{figure}
    \subfloat[mimi2-iaccd]{\includegraphics[width=0.48\linewidth]{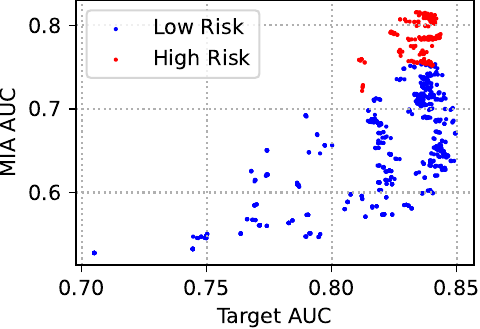}}
    \hfill
    \subfloat[hospital]{\includegraphics[width=0.48\linewidth]{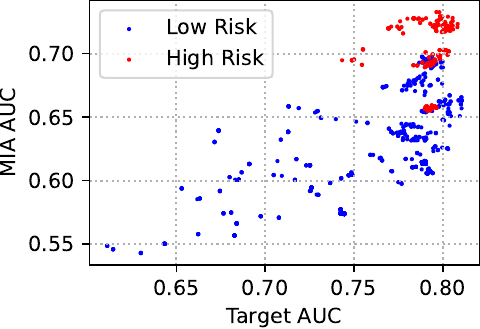}}\\
    \subfloat[indian liver]{\includegraphics[width=0.48\linewidth]{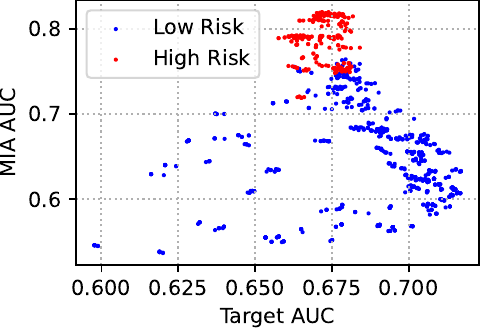}}
    \hfill
    \subfloat[mammography]{\includegraphics[width=0.48\linewidth]{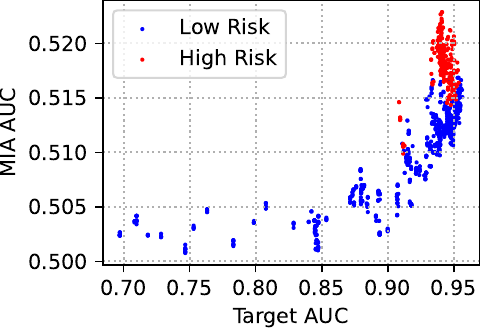}}\\
    \subfloat[sick]{\includegraphics[width=0.48\linewidth]{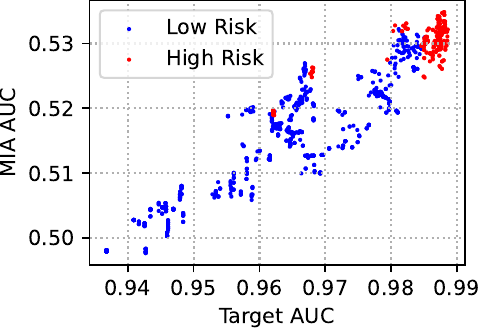}}
    \hfill
    \subfloat[synth-ae]{\includegraphics[width=0.48\linewidth]{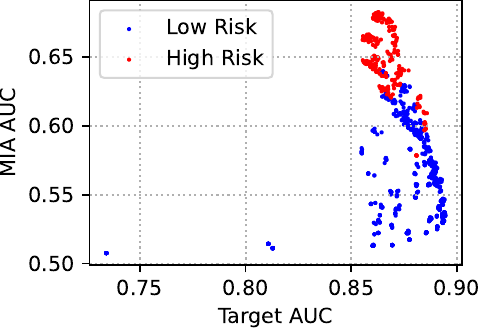}}
    \caption{Privacy risk (MIA AUC) vs.\ accuracy (target AUC) for XGBoost target models. Points in red denote hyperparameter combinations predicted to be in the top 20\% (high risk) most vulnerable to MIA.}%
    \label{fig:scatter_xgboost}
\end{figure}

%% file: figs/figure-4.tex
\begin{figure}[p]
    \centering
    \includegraphics[width=1.0\textwidth]{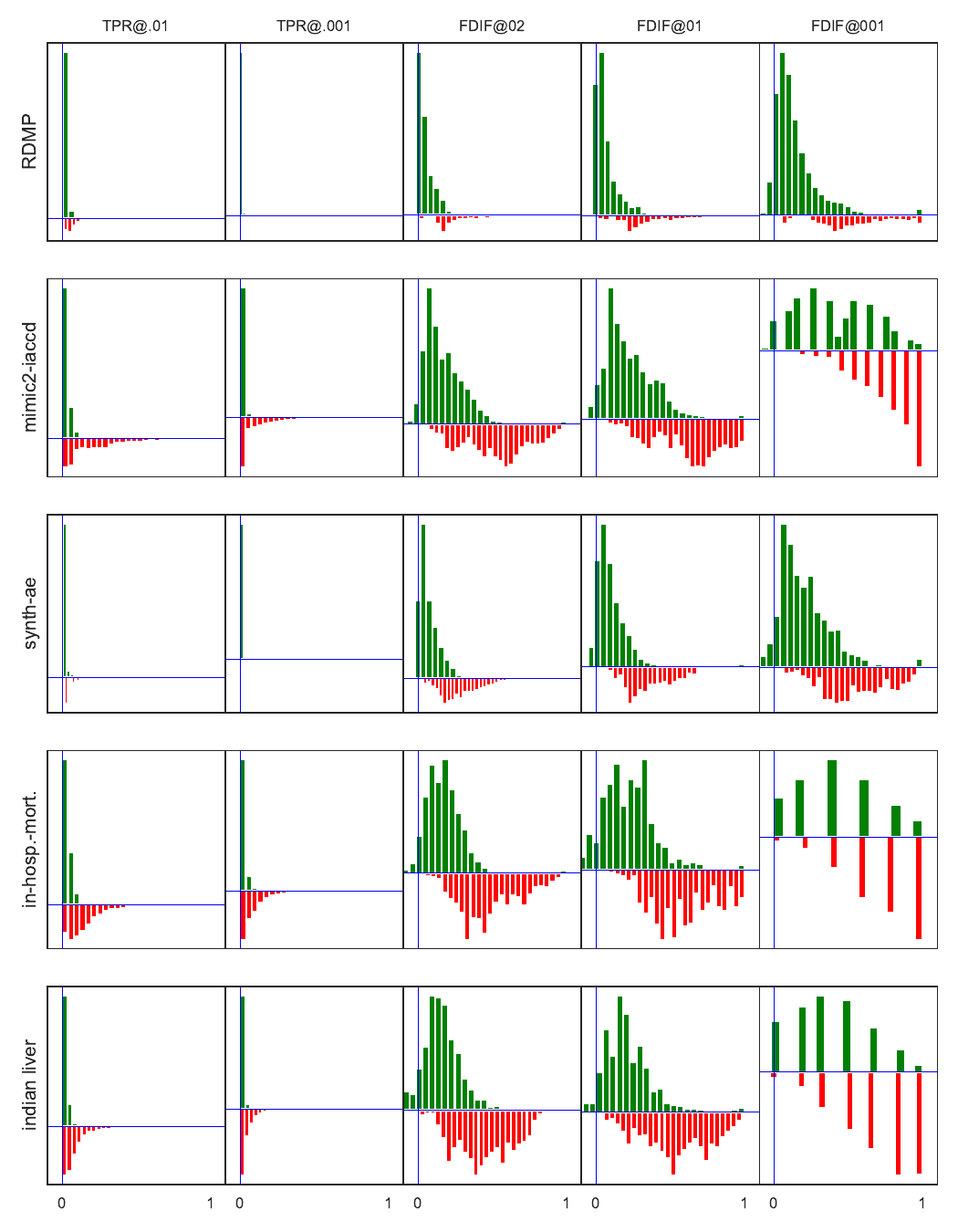}
    \caption{Frequencies of observed risk metrics for different datasets when combined structural risk is 0 (above, green) or 1 (below, red). Note each subplot has different $y$ scales. Blue lines indicate frequency or metric value of 0. Note that while TPR metrics are always $\ge 0$, a poor membership attack model may have FDIF $<0$.}%
    \label{fig:violin-dataset-metric}
\end{figure}

%% file: figs/figure-5.tex
\begin{figure}[t]
    \centering
    \includegraphics[scale=.8]{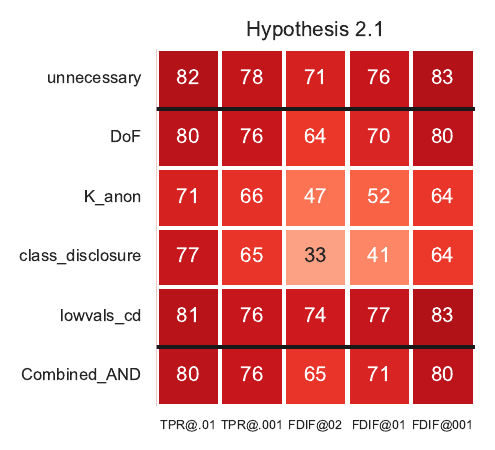}%
    \includegraphics[scale=.8]{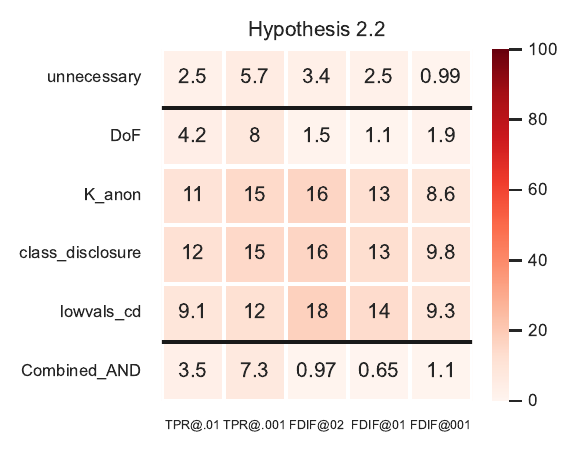}
    \caption{Heatmaps showing percentages of cases providing evidence against the two hypotheses, cross-tabulated by structural (rows) and LiRA (columns) vulnerability metrics.}%
    \label{fig:heatmaps}
\end{figure}

%% file: figs/figure-6.tex
\begin{figure}[t!]
    \centering
    \includegraphics[width=1.0\textwidth]{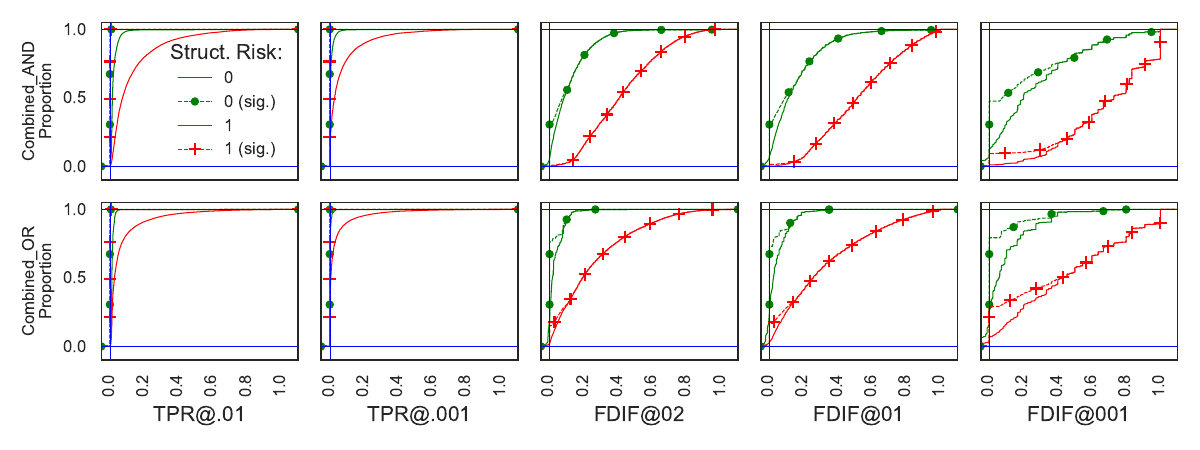}
    \caption{Empirical cumulative distribution functions for proportion of cases ($y$-axis) vs.\ value of different metrics ($x$-axis). Top row shows vulnerability to \textit{all} structural metrics, where red lines correspond to hypothesis 2.2. Bottom row shows vulnerability to \textit{any} of the structural metrics, where green lines relate to hypothesis 2.1. Green lines are when the structural risk metric is 0 (low risk) and red lines when risk is 1 (high risk) with solid lines for all cases, and dashed lines with markers showing only those cases where the reported MIA metric is statistically significant with $>$95\% confidence. In an ideal case, the green lines would rise immediately to 1 at the origin; that is, all models with no structural risk would have zero MIA risk. By contrast, ideally the red lines (structural risk is present) would begin low then rise only if significant MIA metrics were observed. If all observed results were statistically significant then the paired lines (i.e., with/without cross markers) would be identical.}%
    \label{fig:ecdf_metrics_ssd}
\end{figure}

%% file: 05_conclusion.tex
\section{Conclusion}%
\label{sec:conclusion}

This article has presented the first empirical analysis of tree-based classification model hyperparameters in terms of their susceptibility to MIA\@. Surprisingly, regardless of MIA success metric, the rank order correlation of disclosure risk is seen to be preserved across datasets. It has been shown that simple human-interpretable rulesets learned on one dataset can predict with approximately 90--95\% accuracy whether a given hyperparameter combination is one of the 20\% most vulnerable on other unseen datasets. Significantly, it has been shown that for most datasets, high target model accuracy is still obtainable with hyperparameter combinations that are not in this highest risk category.

The use of more complex rules/models such as deep neural networks would likely increase the accuracy of risk prediction, however the use of a simple Decision Tree to extract the ruleset enables easy interpretation by researchers so that they can more easily adjust their preferred hyperparameters into a safer region that maintains sufficient classification accuracy. Future work may explore the use of rule-based learners such as the XCSF learning classifier system~\cite{Preen:2024} to extract interpretable rules with improved accuracy.

While these simple rules offer valuable insights, it is important to acknowledge that they cannot definitively determine `safe' hyperparameter combinations, as absolute risk scores are highly dependent on dataset characteristics such as distribution, imbalances, outliers, and size. Nevertheless, our results suggest that the \textit{relative} risk of hyperparameters can remain surprisingly consistent across diverse datasets. This consistency implies that hyperparameter values associated with the highest risk can reasonably be excluded from the tuning process---and consequently, before running expensive MIAs---without significant loss of performance. Although further dataset testing with the rules extracted here may reveal limitations in generalisation, the existence of dataset classes where rules learned from one dataset successfully generalise to others presents an exciting opportunity to exploit similarities within secure data facilities such as TREs, which often house large numbers of related datasets.

Some datasets/applications may still require some privacy trade-off for sufficient utility, but it seems reasonable that as the first stage of risk assessment it should be incumbent on the researchers wishing to release a model to first show that it is necessary to use these highest risk hyperparameters. In the event that the rules do not appear to be generalising to a particular dataset after sampling some hyperparameter values, then the use of surrogate modelling to efficiently identify the best privacy-utility trade-off could be used (i.e., generalising within the dataset rather than across datasets), however this may not be possible for very large models and datasets.

Once the ML model has been trained, the use of computationally cheap structural metrics as shown here can serve as sufficient, although not necessary, indicators of MIA vulnerability. Again, this approach for identifying vulnerable models cannot certify a model as safe, but potentially enables a large number of high risk models to be filtered and therefore a significant reduction in the number needing to undergo expensive MIAs such as LiRA\@.

Current work is integrating the learned rulesets and structural attacks within the open source SACRO Python toolkit~\cite{Smith:2024,Preen:2025} to provide a comprehensive hierarchical framework to perform these risk assessments, in addition to running current state-of-the-art MIAs. Future work may explore additional target models, hyperparameters, and datasets. Including, for example, the consideration of non-tree models such as neural networks. Potentially, a similar approach could be undertaken by using feature encoding schemes similar to those used for the hyperparameters when surrogate modelling neural architectures, e.g., hand-crafted features such as the number of parameters, depth, width, number of convolutions, or sequence based representations such as tokenising the architecture~\cite{White:2021,Ning:2021}. The structural metrics presented here also have similarities with zero-cost proxies for estimating the accuracy of deep neural networks~\cite{Mellor:2021} and issues such as overparameterising and lack of regularisation hyperparameters may be indicative of MIA risk.

The data produced from experimentation and analysis code used to produce the figures and statistics are available in~\cite{Preen:2025b}.

%% file: 06_acknowledgement.tex
\section*{Acknowledgements}

This work was funded by UK Research and Innovation (Grant Numbers MC\_PC\_23006, MC\_PC\_21033, and MC\_PC\_24038) as part of the Data and Analytics Research Environments UK (DARE UK) programme, delivered in partnership with Health Data Research UK (HDR UK) and Administrative Data Research UK (ADR UK).

%% file: 07_appendix.tex
\appendix
\section{Appendix}

\subsection{Rulesets for Predicting Unnecessary Risk}%
\label{sec:rules}

Figure~\ref{fig:rules_combined} shows the rulesets distilled from Decision Tree classifiers trained to recognise whether a set of hyperparameters will produce a model ranked in the top 20\% (high risk) for MIA risk on the mimic dataset.

\input{figure-7.tex}

%% file: figs/figure-7.tex
\begin{figure}[t]
    \small
    \centering
    \begin{tabular}{l p{0.8\textwidth}}
        \toprule
        \multicolumn{2}{l}{{\bf Decision Tree:}} \\
        Rule 1: & \texttt{{\bf if} (max\_depth $>$ 7.5) and (min\_samples\_leaf $\leq$ 7.5) and (min\_samples\_split $\leq$ 15) {\bf then} high risk}\\
        Rule 2: & \texttt{{\bf if} (splitter $=$ best) and  (max\_depth $>$ 7.5) and (min\_samples\_leaf $\leq$ 7.5) and (min\_samples\_split $>$ 15) {\bf then} high risk}\\
        Rule 3: & \texttt{{\bf if} (splitter $=$ best) and (max\_depth $>$ 7.5) and (7.5 $<$ min\_samples\_leaf $\leq$ 15) and (max\_features $=$ None) {\bf then} high risk}\\
        Rule 4: & \texttt{{\bf if} (splitter $=$ best) and  (3.5 $<$ max\_depth $\leq$ 7.5) and (max\_features $=$ None) and (min\_samples\_leaf $\leq$ 7.5) {\bf then} high risk}\\
        Rule 5: & \texttt{{\bf if} (splitter $=$ random) and (max\_depth $>$ 7.5) and (min\_samples\_leaf $\leq$ 7.5) and (max\_features $=$ None) {\bf then} high risk}\\
        \midrule
        \multicolumn{2}{l}{{\bf Random Forest:}} \\
        Rule 1: & \texttt{{\bf if} (max\_depth $>$ 3.5) and (n\_estimators $>$ 35) and  (max\_features $\neq$ None)  {\bf then} high risk}\\
        Rule 2: & \texttt{{\bf if} (max\_depth $>$ 3.5) and (n\_estimators $>$ 35) and (min\_samples\_split $\leq$ 15) and (max\_features $=$ None) and (bootstrap $=$ True) {\bf then} high risk}\\
        Rule 3 & \texttt{{\bf if} (max\_depth $>$ 7.5) and (15 $<$ n\_estimators $\leq$ 35) and (min\_samples\_leaf $\leq$ 15) and (bootstrap $=$ False) {\bf then} high risk}\\
        \midrule
        \multicolumn{2}{l}{{\bf XGBoost:}} \\
        Rule 1: & \texttt{{\bf if} (max\_depth $>$ 3.5) and (3.5 $<$ n\_estimators $\leq$ 12.5) and  (min\_child\_weight $\leq$ 1.5) {\bf then} high risk}\\
        Rule 2: & \texttt{{\bf if} (max\_depth $>$ 3.5) and (n\_estimators $>$ 12.5) and   (min\_child\_weight $\leq$ 3) {\bf then} high risk}\\
        Rule 3: & \texttt{{\bf if} (max\_depth $>$ 3.5) and (n\_estimators $>$ 62.5) and (3 $<$ min\_child\_weight $\leq$ 6) {\bf then} high risk}\\
        \bottomrule
    \end{tabular}
    \caption{MIA (high/low) risk rules for each target model; extracted from the mimic dataset.}%
    \label{fig:rules_combined}
\end{figure}